\documentclass[conference]{IEEEtran}

\usepackage{amsmath,amssymb}
\usepackage{graphicx}
\usepackage{booktabs}
\usepackage{hyperref}
\usepackage{xcolor}
\usepackage{algorithm}
\usepackage{algorithmic}
\usepackage{multirow}
\usepackage{subcaption}
\usepackage{tikz}
\usepackage{pgfplots}
\pgfplotsset{compat=1.18}
\usetikzlibrary{arrows.meta,positioning,shapes.geometric,fit,calc,backgrounds,patterns}

\title{ARCS: Autoregressive Circuit Synthesis with\\
Topology-Aware Graph Attention and Spec Conditioning}

\author{
\IEEEauthorblockN{Tushar Dhananjay Pathak}
\IEEEauthorblockA{New York University\\
\texttt{tdp9953@nyu.edu}}
}

\begin{document}
\maketitle

\begin{abstract}
This paper presents ARCS (\textbf{A}uto\textbf{r}egressive
\textbf{C}ircuit \textbf{S}ynthesis), a system for \emph{amortized}
analog circuit generation. ARCS produces complete, SPICE-simulatable
designs (topology and component values) in milliseconds rather than
the minutes required by search-based methods. A hybrid pipeline
combines two learned generators, a graph VAE and a flow-matching
model, with SPICE-based ranking. It achieves \textbf{99.9\%
simulation validity} (reward 6.43/8.0) across 32 topologies using
only 8 SPICE evaluations, 40$\times$ fewer than genetic algorithms.
For single-model inference, a topology-aware Graph Transformer with
Best-of-3 candidate selection reaches \textbf{85\% simulation
validity in 97\,ms}, over 600$\times$ faster than random search. The
key technical contribution adapts \emph{Group Relative Policy
Optimization} (GRPO) to multi-topology circuit reinforcement
learning. GRPO resolves a critical failure mode of REINFORCE,
cross-topology reward distribution mismatch, through per-topology
advantage normalization. This improves simulation validity by
+9.6~percentage points over REINFORCE in only 500 RL steps ($10\times$
fewer). \emph{Grammar-constrained decoding} additionally guarantees
100\% structural validity by construction via topology-aware token
masking.
\end{abstract}

\section{Introduction}
\label{sec:intro}

Designing an analog circuit takes days. Engineers select a topology,
choose component values, simulate, and iterate, a cycle repeated
until specifications are met. Recent machine learning approaches
automate parts of this workflow, but each leaves critical gaps.

\textbf{Topology-only generation.} AnalogGenie~\cite{analoggenie}
generates circuit topologies using an autoregressive transformer over
pin-level Eulerian sequences, achieving 93.2\% validity after
proximal policy optimization (PPO) fine-tuning. However, it produces
no component values; a separate genetic algorithm (GA) must size
every generated circuit, adding minutes of computation. It also lacks specification conditioning: circuits are
generated randomly, then filtered post-hoc.

\textbf{Text-based generation.} CircuitSynth~\cite{circuitsynth} and
AutoCircuit-RL~\cite{autocircuitrl} fine-tune large language models
(GPT-Neo, 2.7B parameters) on raw SPICE netlist text. These models
predict characters, not circuit components, leading to syntax errors in
generated netlists and no semantic understanding of component values.

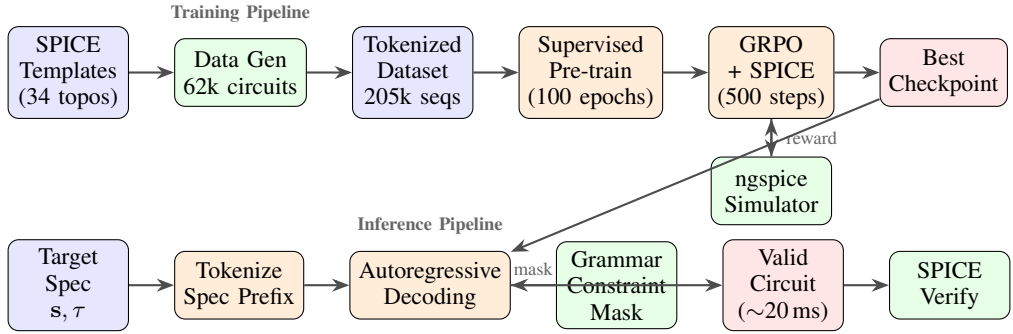
\begin{figure*}[t]
\centering
\begin{tikzpicture}[>=Stealth, node distance=0.6cm,
  block/.style={draw, rounded corners, minimum height=0.9cm,
    minimum width=1.6cm, align=center, font=\small},
  data/.style={block, fill=blue!10},
  model/.style={block, fill=orange!15},
  process/.style={block, fill=green!10},
  result/.style={block, fill=red!10},
  arr/.style={->, thick, color=black!70},
  lbl/.style={font=\scriptsize, color=black!60}
]
\node[data] (templates) {SPICE\\Templates\\(34 topos)};
\node[process, right=of templates] (datagen) {Data Gen\\62k circuits};
\node[data, right=of datagen] (dataset) {Tokenized\\Dataset\\205k seqs};
\node[model, right=of dataset] (sl) {Supervised\\Pre-train\\(100 epochs)};
\node[model, right=of sl] (rl) {GRPO\\+ SPICE\\(500 steps)};
\node[result, right=of rl] (ckpt) {Best\\Checkpoint};

\draw[arr] (templates) -- (datagen);
\draw[arr] (datagen) -- (dataset);
\draw[arr] (dataset) -- (sl);
\draw[arr] (sl) -- (rl);
\draw[arr] (rl) -- (ckpt);

\node[process, below=0.5cm of rl] (spice) {ngspice\\Simulator};
\draw[arr] (rl) -- (spice);
\draw[arr] (spice) -- node[lbl,right,xshift=2pt]{reward} (rl);

\node[data, below=1.6cm of templates] (spec) {Target\\Spec\\$\mathbf{s},\tau$};
\node[model, right=of spec] (encode) {Tokenize\\Spec Prefix};
\node[model, right=of encode] (decode) {Autoregressive\\Decoding};
\node[process, right=of decode] (constrain) {Grammar\\Constraint\\Mask};
\node[result, right=of constrain] (circuit) {Valid\\Circuit\\($\sim$20\,ms)};
\node[process, right=of circuit] (verify) {SPICE\\Verify};

\draw[arr] (spec) -- (encode);
\draw[arr] (encode) -- (decode);
\draw[arr] (constrain) -- node[lbl,above]{mask} (decode);
\draw[arr] (decode) -- (circuit);
\draw[arr] (circuit) -- (verify);
\draw[arr] (ckpt) -- (decode);

\node[lbl, above=0.1cm of datagen] {\textbf{Training Pipeline}};
\node[lbl, above=0.1cm of decode] {\textbf{Inference Pipeline}};
\end{tikzpicture}
\caption{ARCS system overview. \textbf{Top}: Training pipeline. SPICE
templates generate data, supervised pre-training learns the sequence
distribution, and GRPO with SPICE-in-the-loop per-topology advantages
refines value quality. \textbf{Bottom}: Inference pipeline. A target
specification is tokenized, the trained model autoregressively generates
component tokens with grammar-constrained masking, producing a valid
circuit in $\sim$20\,ms.}
\label{fig:overview}
\end{figure*}

\textbf{This work.} ARCS (Figure~\ref{fig:overview}) addresses all three gaps
simultaneously. Given one of 32 parameterized topology templates and a
target specification, ARCS generates complete component-value assignments
in a single forward pass. The three core contributions are:
\begin{enumerate}
    \item \textbf{Group Relative Policy Optimization (GRPO) for
    multi-topology reinforcement learning (RL)}: REINFORCE suffers
    from cross-topology reward
    distribution mismatch: easy topologies dominate gradient
    updates while hard topologies are abandoned. GRPO resolves this
    via per-topology advantage
    normalization (Eq.~\ref{eq:grpo}), achieving 53.1$\pm$3.1\%
    simulation validity (+9.6 percentage points (pp) over REINFORCE) with 10$\times$
    fewer RL steps.
    \item \textbf{Grammar-constrained decoding}: A state-machine-based
    token masking scheme guarantees 100\% structural validity \emph{by
    construction}, without RL training or post-hoc filtering. Combined
    with Best-of-3 candidate selection, this yields 85\% simulation
    validity in 97\,ms.
    \item \textbf{Hybrid multi-source ranking}: A pipeline combining
    two complementary generators (a graph-structured VAE (VCG) and a
    Constrained Circuit Flow Matching model (CCFM)) with SPICE-based
    ranking achieves 99.9\% simulation validity and reward 6.43/8.0
    with only 8 SPICE evaluations (40$\times$ fewer than GA).
\end{enumerate}

The key contribution of ARCS is \emph{amortized inference}: a trained
model that produces reasonable first-shot designs in $\sim$20\,ms,
enabling rapid prototyping and design-space exploration at a cost
$>$1000$\times$ lower than conventional search. ARCS does not yet match
the per-design quality of search baselines (5.48 vs.\ 7.48 reward), but
the paradigms are complementary: ARCS-seeded GA recovers 96.6\%
of cold-start quality with 49\% fewer simulations.

\section{Related Work}
\label{sec:related}

\subsection{Autoregressive Circuit Generation}

AnalogGenie~\cite{analoggenie} introduced Eulerian circuit representations
for autoregressive generation, training a GPT decoder (11.8M parameters)
on 3,502 circuits manually curated from IEEE publications.
After PPO fine-tuning, it achieves 93.2\% structural validity. Key
limitations: no component values in the vocabulary, no specification
conditioning, and reliance on a GA for post-hoc sizing.

CircuitSynth~\cite{circuitsynth} and AutoCircuit-RL~\cite{autocircuitrl}
fine-tune GPT-Neo (2.7B parameters) on SPICE netlist text, treating
circuit generation as a text completion task. While this includes
component values in the output, the model operates at the character
level with no circuit-specific inductive bias.

LaMAGIC~\cite{lamagic} generates analog IC topologies using language
models with graph-based representations. While it introduces
circuit-aware tokenization, it targets transistor-level integrated
circuit (IC) design rather than board-level circuits with discrete
component values.
AnalogXpert~\cite{analogxpert} incorporates circuit design expertise
into large language models for topology synthesis, demonstrating that
domain knowledge can guide LLM-based generation.
AstRL~\cite{astrl} applies deep RL to analog and
mixed-signal circuit synthesis, training an RL agent to select
topologies and size components end-to-end.
INSIGHT~\cite{insight} uses an autoregressive transformer as a
fast surrogate for SPICE simulation, achieving near-simulator
accuracy at orders-of-magnitude lower cost; however, it is a
\emph{predictor} of circuit performance, not a \emph{generator}
of new designs.

\subsection{Circuit Sizing and Optimization}

Several prior works size or score analog circuits without generating
full topologies. AutoCkt~\cite{autockt} and
GCN-RL~\cite{wang2020gcnrl} apply deep RL to transistor sizing; the
latter uses graph neural networks (GNNs) for cross-topology transfer.
Lyu~\emph{et~al.}~\cite{lyu2018weibo} size circuits via Bayesian
optimization with Gaussian process surrogates. Other methods are
topology-limited or non-generative: CktGNN~\cite{cktgnn} handles only
operational amplifiers, Fan~\emph{et~al.}~\cite{fan2024graph} predict
power converter performance but do not generate new designs, and
FALCON~\cite{falcon} combines GNN prediction with layout
co-optimization but cannot synthesize new topologies. Closest to
ARCS, DiffCkt~\cite{diffckt} applies diffusion models to
transistor-level generation, showing that generative approaches
beyond variational autoencoders (VAEs) and autoregressive transformers
can produce valid topologies.

\subsection{Flow-Based Generative Models}

Conditional flow matching (CFM)~\cite{lipman2023flow} learns
continuous-time transport maps between a prior and data distribution
via optimal-transport velocity fields, offering straighter sampling
trajectories than diffusion models and avoiding the KL-constrained
latent space of VAEs. CFM has been applied to molecular
generation~\cite{song2024equivariant} and protein design~\cite{jing2023eigenfold},
but not, to date, to circuit synthesis.
For discrete graph generation, DiGress~\cite{vignac2023digress} applies
discrete denoising diffusion to molecule and planar graphs; however, it
does not handle mixed discrete-continuous outputs (topology + values)
or specification conditioning.
ARCS introduces CCFM (Section~\ref{sec:ccfm}), the first application
of flow matching to analog circuit generation.

\subsection{Key Differences}

\begin{table}[t]
\centering
\caption{Comparison of generative circuit design methods.}
\label{tab:related}
\begin{tabular}{@{}lcccccc@{}}
\toprule
& Values & Specs & Graph & Valid & Speed & Data \\
\midrule
AnalogGenie & \texttimes & \texttimes & \texttimes & 93.2\% & $\sim$1\,s & Manual \\
LaMAGIC     & \texttimes & \texttimes & \checkmark & N/A & N/A & Synthetic \\
CircuitSynth & \checkmark & \texttimes & \texttimes & N/A & $\sim$10\,s & RL only \\
AstRL       & \checkmark & \checkmark & \texttimes & N/A & N/A & RL only \\
DiffCkt     & \checkmark & \texttimes & \checkmark & N/A & N/A & Synthetic \\
\textbf{ARCS} & \checkmark & \checkmark & \checkmark & \textbf{100\%}$^\dagger$ & 0.02\,s & Auto \\
\bottomrule
\multicolumn{7}{@{}l}{\footnotesize $^\dagger$Structural validity with grammar-constrained decoding.}\\
\multicolumn{7}{@{}l}{\footnotesize Sim-validity: 85\% (Best-of-3), 99.9\% (hybrid pipeline).}
\end{tabular}
\end{table}

ARCS is the first system that jointly generates
component values conditioned on both topology and target specifications
while incorporating circuit-graph inductive bias
(Table~\ref{tab:related}). Notably, ARCS performs
\emph{conditioned component sizing} across 32 predefined topology
templates, not topology \emph{discovery}. The topology is selected
from a known library; the model's task is to predict all component
values that satisfy the target specification.
AnalogGenie generates diverse topologies but
outputs only graph structure without component values. ARCS produces
immediately simulatable netlists, closing the gap between generation
and verification.

\section{Method}
\label{sec:method}

\subsection{Problem Formulation}

Given a target specification $\mathbf{s} = (s_1, \ldots, s_K)$ (e.g.,
$V_\text{out}=5$\,V, $I_\text{out}=1$\,A) and a topology label $\tau$
(e.g., ``buck''), generate a sequence of component tokens
$\mathbf{c} = (c_1, v_1, c_2, v_2, \ldots, c_N, v_N)$ where each
$(c_i, v_i)$ pair specifies a component type and its value. The
generated circuit should be electrically valid and meet the target
specification when simulated.

\subsection{Tokenizer}

The tokenizer defines a domain-specific vocabulary of 706 tokens organized into
seven semantic categories:
\begin{itemize}
    \item \textbf{Special tokens} (5): \texttt{START}, \texttt{END},
    \texttt{PAD}, \texttt{SEP}, \texttt{INVALID}.
    \item \textbf{Component tokens} (20): \texttt{MOSFET\_N},
    \texttt{RESISTOR}, \texttt{CAPACITOR}, \texttt{INDUCTOR},
    \texttt{DIODE}, \texttt{OPAMP}, \texttt{TRANSFORMER}, etc.
    \item \textbf{Topology tokens} (40): Slots for 34 named circuit
    topologies plus 6 reserved tokens for future expansion.
    \item \textbf{Spec tokens} (20): \texttt{SPEC\_VIN},
    \texttt{SPEC\_VOUT}, \texttt{SPEC\_IOUT}, \texttt{SPEC\_FSW},
    \texttt{SPEC\_GAIN}, \texttt{SPEC\_BW}, etc.
    \item \textbf{Pin tokens} (21): Component pin labels
    (\texttt{PIN\_DRAIN}, \texttt{PIN\_POS}, etc.).
    \item \textbf{Net/connection tokens} (100): Circuit net identifiers.
    \item \textbf{Value tokens} (500): Log-uniformly spaced bins covering
    $10^{-12}$ to $10^{6}$, providing $\sim$28 bins per decade for
    sub-decade resolution.
\end{itemize}

Each circuit is encoded as a flat sequence:
\[
\begin{aligned}
&\texttt{START} \!\to\! \texttt{TOPO}_\tau \!\to\! \texttt{SEP} \!\to\!
\underbrace{s_1, v_{s_1}, \ldots, s_K, v_{s_K}}_{\text{spec prefix}} \!\to\! \texttt{SEP} \\
&\quad\to\! \underbrace{c_1, v_1, \ldots, c_N, v_N}_{\text{components}} \!\to\! \texttt{END}
\end{aligned}
\]

The log-uniform binning ensures that physically meaningful value ratios
(e.g., $10\,\text{k}\Omega$ vs.\ $1\,\text{k}\Omega$) map to
proportional token distances, and data augmentation via random component
order shuffling ($5\times$) teaches the model that component ordering
is arbitrary.

\begin{figure}[t]
\centering
\small
\begin{tabular}{@{}l@{}}
\toprule
\textbf{Example: Generated Buck Converter} \\
\midrule
\texttt{START TOPO\_BUCK SEP} \\
\texttt{SPEC\_VIN \colorbox{blue!10}{VAL\_12.0} SPEC\_VOUT \colorbox{blue!10}{VAL\_5.0}} \\
\texttt{SPEC\_IOUT \colorbox{blue!10}{VAL\_1.0} SEP} \\
\texttt{INDUCTOR \colorbox{orange!15}{VAL\_100$\mu$H} CAPACITOR \colorbox{orange!15}{VAL\_22$\mu$F}} \\
\texttt{RESISTOR \colorbox{orange!15}{VAL\_5m$\Omega$} MOSFET \colorbox{orange!15}{VAL\_15m$\Omega$}} \\
\texttt{END} \\
\midrule
\footnotesize $\downarrow$ Decoded to SPICE netlist $\downarrow$ \\
\midrule
\texttt{L1 sw\_node vout 1.000e-04} \\
\texttt{C1 cap\_node 0 2.200e-05 IC=5.0} \\
\texttt{S1 input sw\_node pwm 0 SMOD} \\
\texttt{Rload vout 0 5.0} \\
\bottomrule
\end{tabular}
\caption{Example ARCS-generated buck converter. \textbf{Top}: Token
sequence with \colorbox{blue!10}{spec values} and
\colorbox{orange!15}{component values}. \textbf{Bottom}: Decoded SPICE
netlist fragment. The model generates the full sequence in $\sim$20\,ms;
grammar constraints ensure structural validity at each step.}
\label{fig:example}
\end{figure}

\subsection{Model Architecture}
\label{sec:arch}

ARCS explores three progressively more expressive architectures, all
sharing a common GPT-style backbone: hidden dimension
$d_\text{model}\!=\!256$, $n_\text{layers}\!=\!6$ transformer layers,
$n_\text{heads}\!=\!4$ attention heads, feed-forward dimension
$d_\text{ff}\!=\!1024$, SwiGLU activations~\cite{shazeer2020glu},
RMSNorm~\cite{zhang2019root}, and learned token-type embeddings
(7 categories). All models use weight-tied language model heads.

\textbf{(A) Baseline GPT} ($\sim$6.5M parameters). A standard
decoder-only transformer with a single weight-tied output head
over the full vocabulary. This model treats all tokens (structural
and value) identically.

\textbf{(B) Two-Head Model} ($\sim$6.8M parameters; $+$4.7\% over
Baseline GPT). Predicting \emph{which component to place} and
predicting \emph{what value it should have} are fundamentally
different tasks. The two-head model separates these:
\begin{itemize}
    \item A \emph{structure head} (weight-tied with the token
    embedding) predicts component type, topology, spec, and special
    tokens.
    \item A \emph{value head}, a dedicated 2-layer multi-layer
    perceptron (MLP) with SiLU activations
    ($256 \!\to\! 256 \!\to\! 256$, residual connection, linear
    projection to $|V|$), specializes in predicting numerical value
    tokens.
\end{itemize}
At each timestep, the model selects which head to use based on
whether the next token is expected to be a value token
(determined by the preceding component token).

\textbf{(C) Graph Transformer (GT)} ($\sim$6.8M parameters; $+$5.0\%
over Baseline GPT). Circuit components form a graph defined by their
electrical connectivity. The GT injects this structure via
two mechanisms:
\begin{enumerate}
    \item \textbf{Topology-aware attention bias.} For each of the
    32 topologies, ARCS precomputes a binary adjacency matrix
    $\mathbf{A}_\tau \in \{0,1\}^{M \times M}$ where $M$ is the
    maximum number of components. During self-attention, a
    learned scalar bias $b_\tau^{(l)}$ augments the attention logits for
    positions $(i,j)$ where $A_{\tau,ij}=1$:
    \begin{equation}
    \text{Attn}(Q,K)_{ij} = \frac{Q_i K_j^\top}{\sqrt{d_k}} +
    b_\tau^{(l)} \cdot A_{\tau,ij},
    \label{eq:graph_attn}
    \end{equation}
    This gives the model a soft prior that electrically connected
    components should attend to each other more strongly.
    \item \textbf{Random-walk positional encoding (RWPE).} Each
    component receives a $K$-dimensional positional feature derived
    from the topology's random-walk transition matrix
    $\mathbf{T} = \mathbf{D}^{-1}\mathbf{A}$, where $\mathbf{D}$
    is the degree matrix. For node~$i$, the RWPE is the vector of
    return probabilities after $k=1,\ldots,K$ steps:
    \begin{equation}
    \text{RWPE}_i = \bigl[\mathbf{T}^1_{ii},\;
    \mathbf{T}^2_{ii},\; \ldots,\; \mathbf{T}^K_{ii}\bigr],
    \label{eq:rwpe}
    \end{equation}
    with $K\!=\!8$ walk lengths. These features are precomputed
    per topology and projected to $d_\text{model}$ via a learned
    two-layer MLP ($8 \!\to\! 64 \!\to\! 256$, Gaussian error linear
    unit (GELU) activation).
    The following value token shares the same RWPE as its parent
    component. Unlike ordinal position indices, RWPE encodes each
    component's \emph{structural role} in the circuit graph; for example,
    high-degree hub nodes (like the switch node in a buck converter)
    have higher return probabilities than peripheral nodes (like the
    load resistor).
\end{enumerate}
The GT retains the two-head output structure from
model~(B), adding 17,648 parameters total: 17,216 for the RWPE
projection MLP and 432 for graph attention biases (per-head
adjacency scalars and edge-type embeddings).

\subsection{Automated Data Generation}
\label{sec:datagen}

Unlike AnalogGenie's~\cite{analoggenie} manually curated dataset
(3,502 circuits hand-collected from IEEE publications), ARCS uses a
fully automated pipeline:
\begin{enumerate}
    \item \textbf{Template definition}: Parameterized SPICE netlist
    templates with physical component bounds (E-series snapped) for
    each of 32 topologies.
    \item \textbf{Random sampling}: Component values drawn
    log-uniformly within bounds.
    \item \textbf{SPICE simulation}: ngspice transient/AC simulation
    with automatic metric extraction (efficiency, output accuracy,
    gain, bandwidth, etc.).
    \item \textbf{Filtering}: Both valid and invalid designs are
    stored; the model learns to avoid failure modes.
\end{enumerate}

This pipeline generated 62,000 circuits (41,064 valid after simulation)
across 34 topologies. After $5\times$ augmentation via component-order
shuffling, the training set contains $\sim$205,000 sequences.

\subsection{Training}

\textbf{Stage 1: Supervised learning (SL) pre-training.}
Next-token prediction with cross-entropy loss, applying $5\times$
weight to value tokens to compensate for their larger vocabulary:
\begin{equation}
\mathcal{L}_\text{SL} = -\sum_t w_t \log p_\theta(x_t \mid x_{<t}),
\label{eq:sl_loss}
\end{equation}
where $w_t = 5$ for value tokens and $w_t = 1$ otherwise.
Training: 100 epochs, batch size 64, AdamW ($\beta_1\!=\!0.9$,
$\beta_2\!=\!0.95$), cosine LR schedule ($3 \times 10^{-4}$
peak) with 5-epoch linear warmup.

\textbf{Stage 2: REINFORCE~\cite{williams1992reinforce} with SPICE reward.}
Stage~2 fine-tunes the best supervised checkpoint with policy gradient
optimization, using SPICE simulation as an oracle reward signal:
\begin{equation}
\nabla_\theta J = \mathbb{E}\!\left[ (R - b) \sum_t \nabla_\theta \log
p_\theta(x_t \mid x_{<t}) \right] - \beta \,
\text{KL}(p_\theta \| p_\text{ref}),
\label{eq:rl_loss}
\end{equation}
where $b$ is an exponential moving average baseline and $\beta$ is an
adaptive KL coefficient~\cite{autocircuitrl} with
$\text{KL}_\text{target}\!=\!0.5$. The reward function $R$ (max~8.0)
decomposes as:
\begin{equation}
R = \underbrace{r_\text{struct}}_{\leq 1.0} +
\underbrace{r_\text{sim}}_{\leq 1.0} +
\underbrace{r_\text{accuracy}}_{\leq 3.0} +
\underbrace{r_\text{efficiency}}_{\leq 2.0} +
\underbrace{r_\text{quality}}_{\leq 1.0}.
\label{eq:reward}
\end{equation}
The sub-rewards are domain-specific and \emph{spec-aware}: power
converters use output voltage error ($|V_\text{out} - V_\text{target}|$)
and efficiency; amplifiers use gain accuracy vs.\ target
($|G_\text{actual} - G_\text{target}|$) and bandwidth; filters use
cutoff frequency accuracy ($|f_{-3\text{dB}} - f_\text{target}| / f_\text{target}$)
and passband gain; oscillators use frequency accuracy
($|f_\text{osc} - f_\text{target}| / f_\text{target}$) and amplitude.
This ensures the reward measures how well generated circuits meet
their target specifications, not just functional correctness.
RL training runs for 5,000 steps with batch size 8, learning rate
$10^{-5}$, temperature 0.8, and top-$k$ 50, requiring $\sim$15 hours
on an Apple M3 with Metal Performance Shaders (MPS) acceleration.

\textbf{Stage 3: Group Relative Policy Optimization (GRPO).}
\label{sec:grpo}
Stage~2 reveals a critical failure mode: global-baseline
REINFORCE suffers from \emph{cross-topology reward distribution
mismatch}. Power converters have max reward $\sim$8.0 while amplifiers
reach $\sim$4.0; when the global baseline $b$ settles near the
cross-topology mean, easy topologies receive systematically positive
advantages while hard topologies receive negative ones, causing the
policy to abandon difficult circuits entirely (Buck: 100\%$\to$10\%
sim-valid under REINFORCE).

GRPO resolves this by computing advantages \emph{within} topology
groups, inspired by the group-relative advantage estimation of
Shao~\emph{et~al.}~\cite{shao2024grpo} but applied per-topology
rather than per-prompt.
Each RL step samples $K$ topologies and generates
$G$ circuits per topology (a ``group''). Advantages are z-scored
per group:
\begin{equation}
\hat{A}_i^{(\tau)} = \frac{R_i - \mu_\tau}{\sigma_\tau + \epsilon},
\label{eq:grpo}
\end{equation}
where $\mu_\tau, \sigma_\tau$ are the mean and standard deviation
of rewards within topology $\tau$'s group. This ensures that every
topology contributes meaningful gradient signal regardless of its
absolute reward scale. The policy gradient becomes:
\begin{equation}
\nabla_\theta J = \sum_\tau \sum_{i \in \text{group}_\tau}
\hat{A}_i^{(\tau)} \sum_t \nabla_\theta \log p_\theta(x_t^{(i)})
- \beta\,\text{KL}(p_\theta \| p_\text{ref}).
\end{equation}
The training uses $K\!=\!3$ topologies and $G\!=\!4$ circuits per step.
With only 500 GRPO steps ($\sim$1 hour, 6,000 SPICE simulations),
the model achieves 53.1$\pm$3.1\% simulation validity, surpassing both
supervised learning (45.4$\pm$3.0\%) and 5,000-step REINFORCE
(43.5$\pm$1.2\%), validating the hypothesis that per-topology normalization
is essential for multi-topology RL.

\subsection{Constrained Circuit Flow Matching (CCFM)}
\label{sec:ccfm}

While the autoregressive approach generates circuits token-by-token,
ARCS additionally includes a \emph{non-autoregressive} generative model
that generates all circuit components in parallel via conditional
flow matching~\cite{lipman2023flow}.

\textbf{Architecture.} CCFM operates in the latent space of a
pre-trained ValidCircuitGen (VCG) encoder, a graph-structured VAE
(4.0M parameters) trained on 41,064 valid circuits with 5
differentiable structural constraints (type coherence, adjacency
symmetry, graph connectivity, degree bounds, value range). The
flow network (3.7M parameters) consists of 4 DiT-style transformer
blocks~\cite{peebles2023dit} with adaptive layer normalization
conditioned on both time embedding $t$ and specification embedding
$\mathbf{s}$:
\begin{equation}
\mathbf{h}^{(l)} = \text{DiT-Block}^{(l)}\!\left(\mathbf{h}^{(l-1)},\;
\gamma(t) + \phi(\mathbf{s})\right),
\end{equation}
where $\gamma(t)$ is a sinusoidal time embedding and $\phi(\mathbf{s})$
is a learned spec projection. Each DiT block applies adaptive LayerNorm
followed by multi-head self-attention and a SwiGLU feed-forward
network, with all conditioning applied via scale-and-shift modulation.

\textbf{Training.} CCFM uses optimal-transport conditional flow
matching~\cite{lipman2023flow}. Given paired samples
$\mathbf{x}_0 \sim \mathcal{N}(0,I)$ and $\mathbf{x}_1$ from the
VCG latent space, the interpolant is $\mathbf{x}_t = (1-t)\mathbf{x}_0
+ t\mathbf{x}_1$ and the target velocity is $\mathbf{v}^* =
\mathbf{x}_1 - \mathbf{x}_0$. The loss is:
\begin{equation}
\mathcal{L}_{\text{CFM}} = \mathbb{E}_{t,\mathbf{x}_0,\mathbf{x}_1}
\!\left[\|\mathbf{v}_\theta(\mathbf{x}_t, t, \mathbf{s}) -
\mathbf{v}^*\|^2\right].
\end{equation}
The VCG encoder is frozen during flow matching training. After
100 epochs of training (25\,s per epoch on MPS), CCFM reduces
validation loss from an initial 0.68 to 0.14.

\textbf{Constraint-guided sampling.} During ordinary differential
equation (ODE) integration
(Euler method, 50 steps), CCFM projects the velocity toward the feasible
circuit set using learned per-constraint guidance weights
$\{w_c\}_{c=1}^5$, initialized to zero and trained end-to-end:
\begin{equation}
\mathbf{v}'_t = \mathbf{v}_\theta(\mathbf{x}_t, t, \mathbf{s}) +
\alpha(t) \sum_c w_c \nabla_{\mathbf{x}_t} C_c(\mathbf{x}_t),
\end{equation}
where $\alpha(t) = 1 - t$ applies stronger guidance early (when
samples are noisy) and relaxes as the trajectory approaches the
data distribution. The five constraint functions $C_c$ are the same
five used by VCG: type coherence, adjacency symmetry, graph
connectivity, degree bounds, and value range.

\subsection{Grammar-Constrained Decoding}
\label{sec:constrained}

A fundamental limitation of unconstrained autoregressive sampling is
that the model may generate structurally invalid token sequences
(e.g., two consecutive component tokens without an intervening value,
or a wrong component type for the target topology). While RL can
improve structural validity to $\sim$100\% (Section~\ref{sec:results}),
it does so statistically, without formal guarantees.

ARCS introduces \emph{grammar-constrained decoding}, which applies a
topology-aware token mask at each autoregressive step, restricting
the sampling distribution to only structurally valid continuations.
This provides 100\% structural validity \emph{by construction},
without RL training or post-hoc filtering.

\begin{figure}[t]
\centering
\begin{tikzpicture}[>=Stealth, node distance=2.4cm,
  state/.style={draw, circle, minimum size=1.4cm, align=center,
    font=\small, thick},
  every edge/.style={draw, thick, ->, >=Stealth}
]
\node[state, fill=blue!12] (ec) {\texttt{EXPECT}\\\texttt{COMP}};
\node[state, fill=orange!15, right=of ec] (ev) {\texttt{EXPECT}\\\texttt{VAL}};
\node[state, fill=green!15, below=1.5cm of $(ec)!0.5!(ev)$] (end) {\texttt{END}};

\draw[->] (ec) to[bend left=25] node[above, font=\scriptsize] {emit component} (ev);
\draw[->] (ev) to[bend left=25] node[below, font=\scriptsize] {emit value} (ec);
\draw[->] (ec) -- node[left, font=\scriptsize, xshift=-2pt] {emit \texttt{END}} (end);

\node[font=\scriptsize, color=blue!70, below=0.05cm of ec, yshift=-0.6cm, align=center]
  {mask: component\\\& \texttt{END} only};
\node[font=\scriptsize, color=orange!70, above=0.05cm of ev, yshift=0.3cm, align=center]
  {mask: value\\tokens only};
\end{tikzpicture}
\caption{Grammar state machine for constrained decoding. At each
autoregressive step, the current state determines which token types
are valid, and a mask zeroes out all others before sampling.
\textsc{Topology}-level constraints further restrict the allowed
component types; \textsc{Full} constraints additionally restrict
value ranges.}
\label{fig:fsm}
\end{figure}
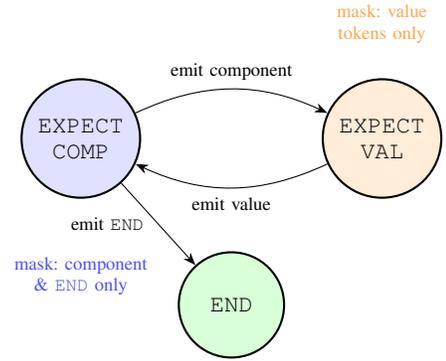

\textbf{Grammar state machine.} The decoder maintains a finite-state machine
with two alternating phases:
\texttt{EXPECT\_COMP} (the next token must be a valid component type
or the \texttt{END} token) and \texttt{EXPECT\_VAL} (the next token
must be a value token). The state machine transitions deterministically
based on the last emitted token.
Figure~\ref{fig:fsm} illustrates the state transitions.

\textbf{Constraint levels.} ARCS defines three levels of increasing
strictness:
\begin{enumerate}
    \item \textsc{Grammar}: Enforces the component--value alternation
    pattern. At \texttt{EXPECT\_COMP} steps, only component and special
    tokens are allowed; at \texttt{EXPECT\_VAL} steps, only value tokens.
    \item \textsc{Topology}: Extends \textsc{Grammar} by restricting
    the set of allowed component types to exactly those required by the
    target topology (e.g., a buck converter requires \texttt{INDUCTOR},
    \texttt{CAPACITOR}, \texttt{DIODE}, \texttt{MOSFET\_N}). This
    ensures correct component types and counts.
    \item \textsc{Full}: Extends \textsc{Topology} by restricting value
    tokens to the physically valid range for each component type
    (e.g., inductors in [1\,$\mu$H, 10\,mH], capacitors in
    [1\,pF, 10\,mF]).
\end{enumerate}

\textbf{Masked sampling.} At each step $t$, the decoder applies a binary mask
$\mathbf{m}_t \in \{0, 1\}^{|V|}$ over the vocabulary
before the softmax:
\begin{equation}
p_\theta(x_t \mid x_{<t}) = \text{softmax}\!\left(
\frac{\mathbf{z}_t + \log \mathbf{m}_t}{\tau}\right),
\label{eq:masked}
\end{equation}
where $\mathbf{z}_t$ are the raw logits and $\log 0 = -\infty$
zeroes out invalid tokens. Since the mask is a deterministic
function of the grammar state and topology, it adds negligible
computational overhead (pre-computed token sets, $<$1\,ms per
sequence). In practice, constrained decoding is actually
\emph{faster} than unconstrained sampling because the model
converges to valid \texttt{END} tokens more quickly, eliminating
wasted tokens from invalid continuations.

\textbf{Lagrangian constraint loss.} For training with constraint
awareness, the model additionally optimizes a differentiable Lagrangian loss:
\begin{equation}
\mathcal{L}_\text{constr} = \sum_{i} \lambda_i \cdot g_i(\theta),
\label{eq:lagrangian}
\end{equation}
where $g_i(\theta)$ are constraint violation penalties (structural
validity, component correctness, value range adherence) and
$\lambda_i$ are adaptive Lagrange multipliers updated via dual
gradient ascent. This allows training to internalize the constraints,
improving generation quality even when decoding without masks.

\subsection{Learned Reward Model}
\label{sec:method-reward}

To improve Best-of-$N$ candidate ranking beyond model confidence,
ARCS also trains a \emph{learned reward model}, a bidirectional transformer
encoder that predicts SPICE simulation reward from token sequences,
following the verifier approach of Cobbe~\emph{et~al.}~\cite{cobbe2021verifiers}.
The architecture uses 2 encoder layers with 128-dim hidden states,
4 attention heads, and a SwiGLU feed-forward network (FFN), followed
by mean-pooling and a 2-layer MLP producing a scalar reward prediction.
Token embeddings are warm-started from the generator via
singular value decomposition (SVD) based projection when the
embedding dimensions differ. The 666K-parameter model is trained on
41K circuits from the automated data pipeline using Huber
loss~\cite{huber1964}.

\section{Experimental Setup}
\label{sec:experiments}

\subsection{Circuit Topologies}

ARCS defines 34 circuit topologies across three tiers, of which 32 are
evaluated in the main experiments:
\begin{itemize}
    \item \textbf{Tier 1} (7 power converters; 5 evaluated): Buck, Boost,
    Buck-Boost, Cuk, SEPIC.
    \item \textbf{Tier 2} (9 signal circuits): Inverting/non-inverting/
    instrumentation/differential amplifiers, Sallen-Key low-pass, high-pass, and band-pass filters,
    Wien bridge and Colpitts oscillators.
    \item \textbf{Tier 2b} (18 extended circuits): BJT amplifiers
    (common emitter/collector/base, cascode), current sources, regulators,
    additional oscillators (Hartley, phase shift), filters (state variable,
    twin-T notch), and extended power (half bridge, push pull, charge pump,
    voltage doubler, zeta converter).
\end{itemize}

\subsection{Baselines}

\textbf{Random Search (RS)}: For each target spec, uniformly sample
200 parameter vectors (log-uniform within bounds), simulate each, and
return the highest-reward design. Total: 200 SPICE simulations per
spec.

\textbf{Genetic Algorithm}: BLX-$\alpha$ crossover in log-space,
Gaussian mutation, tournament selection. Table~\ref{tab:main} uses population 30 over 20 generations ($\sim$630 evals).
The statistical evaluation (Table~\ref{tab:pub_eval}) uses 20 repeats
with population 16 ($\sim$320 evals per spec) to match the compute
budget of the hybrid pipeline.

Both baselines have a fundamental advantage: they search directly in
the known parameter space by evaluating hundreds of value combinations,
while ARCS produces values in a single forward pass from a spec prefix.
Note that ARCS \emph{is} given the correct topology token in its input
prefix (making it a topology-conditioned generation task, not topology
discovery), so the information asymmetry lies in the \emph{search budget},
not topology access. This makes the comparison meaningful: given the same
topology, can a learned model's single-shot prediction compete with
iterative search?

\subsection{Evaluation Protocol}

Each method is evaluated on 160 conditioned test specs (10 per
topology). Metrics:
\begin{itemize}
    \item \textbf{Structural validity}: Well-formed token sequence
    that decodes to a valid circuit.
    \item \textbf{Sim.\ success}: SPICE convergence rate.
    \item \textbf{Sim.\ validity}: Physically plausible simulation
    results (no negative efficiency, reasonable output).
    \item \textbf{Reward}: Domain-specific score (max 8.0), combining
    accuracy, efficiency, and quality metrics.
    \item \textbf{Wall time}: End-to-end time per design including
    SPICE simulation for baselines.
\end{itemize}

\section{Results}
\label{sec:results}

\subsection{Architecture Comparison and Main Results}

\begin{table}[t]
\centering
\caption{Unified comparison across all methods (160 conditioned specs,
10 per topology; autoregressive results averaged over 5 seeds with
standard deviation). Search baselines have oracle access to the correct
topology; ARCS generates from scratch.}
\label{tab:main}
\begin{tabular}{@{}lcccc@{}}
\toprule
Method & Params & Struct & SimValid & Reward \\
\midrule
Random Search    & --- & --- & 81.2\% & 7.28 \\
Genetic Alg.     & --- & --- & 80.0\% & 7.48 \\
\midrule
Baseline GPT (SL)     & 6.5M & 86.0{\tiny$\pm$1.9}\% & 40.1{\tiny$\pm$0.7}\% & 3.37{\tiny$\pm$0.06} \\
Two-Head (SL)         & 6.8M & 96.2{\tiny$\pm$1.0}\% & 50.4{\tiny$\pm$2.2}\% & 3.89{\tiny$\pm$0.09} \\
Graph Trans.\ (SL)    & 6.8M & 93.2{\tiny$\pm$1.4}\% & 45.4{\tiny$\pm$3.0}\% & 3.86{\tiny$\pm$0.15} \\
REINFORCE (5000)      & 6.5M & 95.5{\tiny$\pm$1.3}\% & 43.5{\tiny$\pm$1.2}\% & 3.74{\tiny$\pm$0.04} \\
\textbf{GT+GRPO (500)}  & \textbf{6.8M} & \textbf{96.6{\tiny$\pm$0.5}\%} & \textbf{53.1{\tiny$\pm$3.1}\%} & \textbf{4.15{\tiny$\pm$0.08}} \\
GT+GRPO (3500)        & 6.8M & 92.9{\tiny$\pm$1.3}\% & 48.8{\tiny$\pm$2.2}\% & 3.79{\tiny$\pm$0.05} \\
\midrule
VCG (graph VAE)   & 4.0M & \multicolumn{2}{c}{100\% valid} & --- \\
CCFM (flow match) & 7.7M & \multicolumn{2}{c}{100\% valid} & --- \\
\bottomrule
\end{tabular}
\end{table}

Table~\ref{tab:main} shows the unified comparison averaged over 5
random seeds.
\textbf{Architecture matters.} The two-head architecture improves
simulation validity by $+10.3$\,pp over the baseline by decoupling
structural and numerical prediction. The GT achieves
$+5.3$\,pp over baseline via topology-aware attention biases
(Eq.~\ref{eq:graph_attn}). Notably, the two-head model (50.4\%) outperforms
the GT under supervised learning alone (45.4\%): the GT's
relational inductive bias is not fully exploited by cross-entropy training
on fixed data, but becomes decisive when combined with GRPO
RL, where GT+GRPO (53.1\%) surpasses both.

\begin{figure}[t]
\centering
\begin{tikzpicture}
\begin{axis}[
  ybar,
  bar width=12pt,
  width=\columnwidth,
  height=4.8cm,
  ylabel={Simulation Validity (\%)},
  symbolic x coords={Baseline,Two-Head,GT SL,GT+RL,GT+GRPO,GT+Constr},
  xtick=data,
  x tick label style={font=\scriptsize, rotate=20, anchor=east},
  ymin=0, ymax=110,
  ytick={0,20,40,60,80,100},
  nodes near coords,
  nodes near coords style={font=\tiny},
  every node near coord/.append style={anchor=south},
  enlarge x limits=0.12,
  legend style={at={(0.02,0.98)}, anchor=north west, font=\scriptsize},
  grid=major,
  grid style={dashed, gray!30},
]
\addplot[fill=blue!30] coordinates
  {(Baseline,40.1) (Two-Head,50.4) (GT SL,45.4) (GT+RL,43.5) (GT+GRPO,53.1) (GT+Constr,100)};
\addplot[fill=orange!40] coordinates
  {(Baseline,86.0) (Two-Head,96.2) (GT SL,93.2) (GT+RL,95.5) (GT+GRPO,96.6) (GT+Constr,100)};
\legend{Sim Valid, Structural}
\end{axis}
\end{tikzpicture}
\caption{Simulation and structural validity across model variants
(mean of 5 seeds). GRPO achieves the best sim-validity
among autoregressive methods ($+9.6$\,pp over REINFORCE). Grammar-constrained
decoding (GT+Constr) achieves 100\% on both metrics.}
\label{fig:validity}
\end{figure}
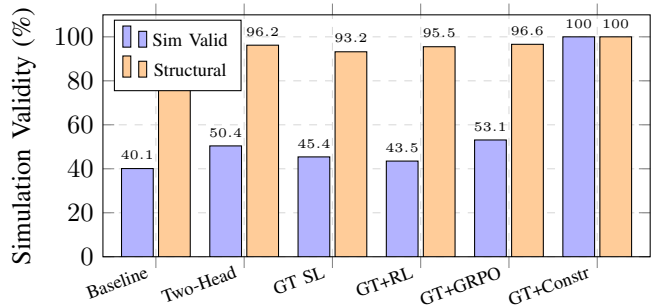

\textbf{GRPO resolves the RL regression.} REINFORCE slightly degrades
simulation validity ($-1.9$\,pp vs.\ SL) due to cross-topology reward
mismatch, while GRPO improves it by $+7.7$\,pp over GT SL and
$+9.6$\,pp over REINFORCE. These differences exceed the $\pm$3\,pp
standard deviations, confirming significance across seeds.
GRPO uses only 500 RL steps
($\sim$6,000 SPICE simulations, $\sim$1 hour) compared to
REINFORCE's 5,000 steps ($\sim$40,000 simulations, $\sim$15 hours)
, a 10$\times$ reduction in training cost. Extended GRPO training
(3,500 total steps) shows diminishing returns: sim-validity drops
to 48.8$\pm$2.2\%, suggesting that 500 steps is the optimal
early-stopping point for this model scale.
This confirms the hypothesis
(Section~\ref{sec:grpo}) that per-topology advantage normalization
is essential for multi-topology RL.

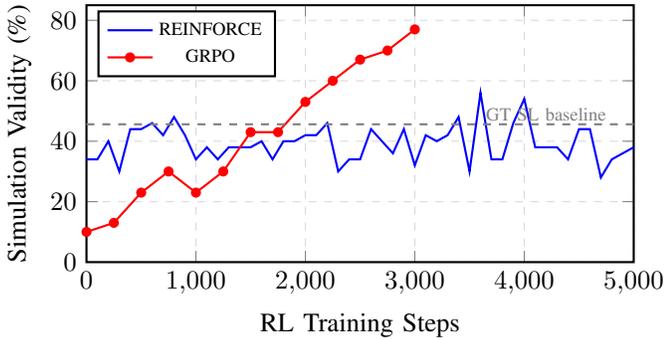
\begin{figure}[t]
\centering
\begin{tikzpicture}
\begin{axis}[
  width=\columnwidth, height=5cm,
  xlabel={RL Training Steps},
  ylabel={Simulation Validity (\%)},
  xmin=0, xmax=5000,
  ymin=0, ymax=85,
  ytick={0,20,40,60,80},
  legend style={at={(0.02,0.98)}, anchor=north west, font=\scriptsize},
  grid=major,
  grid style={dashed, gray!30},
  thick,
]
\addplot[blue, mark=none, thick] coordinates {
  (0,34) (100,34) (200,40) (300,30) (400,44) (500,44) (600,46)
  (700,42) (800,48) (900,42) (1000,34) (1100,38) (1200,34) (1300,38)
  (1400,38) (1500,38) (1600,40) (1700,34) (1800,40) (1900,40)
  (2000,42) (2100,42) (2200,46) (2300,30) (2400,34) (2500,34)
  (2600,44) (2700,40) (2800,36) (2900,44) (3000,32) (3100,42)
  (3200,40) (3300,42) (3400,48) (3500,30) (3600,56) (3700,34)
  (3800,34) (3900,46) (4000,54) (4100,38) (4200,38) (4300,38)
  (4400,34) (4500,44) (4600,44) (4700,28) (4800,34) (4900,36) (5000,38)
};
\addplot[red, mark=*, mark size=1.5pt, thick] coordinates {
  (0,10) (250,13) (500,23) (750,30) (1000,23) (1250,30) (1500,43)
  (1750,43) (2000,53) (2250,60) (2500,67) (2750,70) (3000,77)
};
\addplot[dashed, black!50, thick] coordinates {(0,45.6) (5000,45.6)};
\node[font=\scriptsize, black!50] at (axis cs:4200,49) {GT SL baseline};
\legend{REINFORCE, GRPO}
\end{axis}
\end{tikzpicture}
\caption{RL training dynamics (in-training validation; Table~\ref{tab:main}
reports held-out results). REINFORCE fluctuates around the SL baseline
(dashed). GRPO steadily improves; the best checkpoint (step 500,
53.1\% held-out) was selected via early stopping.}
\label{fig:rl_curves}
\end{figure}

\textbf{Non-autoregressive alternatives.} VCG and CCFM both achieve
100\% structural validity via differentiable constraints, reaching
85.3\% simulation validity (116/136) across 34 topologies. CCFM
trains $\sim$3$\times$ faster than VCG (25\,s vs.\ 71\,s per epoch)
while matching validity.

\subsection{Hybrid Multi-Source Ranking}

\begin{table}[t]
\centering
\caption{Hybrid evaluation with SPICE-based ranking across VCG and CCFM
sources (4 candidates/source; all 34 topologies).
Hybrid picks the highest-reward candidate per topology. Note: the
statistical evaluation in Table~\ref{tab:pub_eval} uses 50 samples on
32 topologies for a more rigorous comparison with baselines.}
\label{tab:hybrid}
\begin{tabular}{@{}lcccc@{}}
\toprule
Method & Struct & SimSuccess & SimValid & Reward \\
\midrule
VCG-only    & 100\% & 95.6\% & 85.3\% & 5.53 \\
CCFM-only   & 100\% & 97.1\% & 85.3\% & 5.50 \\
\textbf{Hybrid (VCG+CCFM)} & \textbf{100\%} & \textbf{97.1\%} & \textbf{94.1\%} & \textbf{6.37} \\
\bottomrule
\end{tabular}
\end{table}

Multi-source ranking is strongly complementary: selecting by SPICE
reward improves average reward by +0.84 over VCG-only and +0.87 over
CCFM-only.

\subsection{Statistical Evaluation with Baselines}

\begin{table}[t]
\centering
\caption{Publication evaluation with spec-aware reward (50 samples per
topology, 32 topologies, bootstrap 95\% CI). Reward measures both
functional correctness and spec compliance (gain accuracy for amplifiers,
cutoff accuracy for filters, frequency accuracy for oscillators).
All pairwise comparisons significant at $p < 0.001$ (Wilcoxon signed-rank).}
\label{tab:pub_eval}
\begin{tabular}{@{}lcccc@{}}
\toprule
Method & SPICE Evals & Reward (95\% CI) & SimValid \\
\midrule
Random Search   & 1   & 5.18 [5.05, 5.31] & 91.4\% \\
Genetic Alg.    & 320 & 7.56 [7.53, 7.60] & 100.0\% \\
\midrule
VCG only        & 1   & 5.34 [5.27, 5.42] & 94.0\% \\
CCFM only       & 1   & 5.51 [5.44, 5.58] & 95.7\% \\
\textbf{Hybrid} & \textbf{8} & \textbf{6.43 [6.38, 6.48]} & \textbf{99.9\%} \\
\bottomrule
\end{tabular}
\end{table}

At equal compute (1 SPICE evaluation), both VCG and CCFM outperform
random sampling. The hybrid pipeline with 8 candidates achieves 99.9\%
simulation validity and reward 6.43 using \textbf{40$\times$ fewer
SPICE evaluations} than GA (7.56). Each pipeline component contributes
significant improvements ($p < 0.001$): VCG (5.34) $\rightarrow$
+CCFM (5.51) $\rightarrow$ +hybrid ranking (6.43).

\subsection{Ablation Studies}

\begin{table}[t]
\centering
\caption{Ablation study (160 samples, baseline model).}
\label{tab:ablation}
\begin{tabular}{@{}lccc@{}}
\toprule
Variant & Struct & SimValid & Reward \\
\midrule
ARCS + RL (full)       & 98.1\% & 52.5\% & 3.49 \\
No RL (supervised only) & 90.6\% & 46.9\% & 3.24 \\
No spec conditioning   & 58.8\% & 38.8\% & 2.60 \\
Tier~1 only (7 topos)  & 100\%  & 20.6\% & 3.77 \\
\bottomrule
\end{tabular}
\end{table}

Removing spec conditioning drops reward by $-25.5\%$ and structural
validity to 58.8\%, confirming that the specification prefix is
essential. RL improves reward by $+0.25$ and validity by $+5.6$\,pp.
The Tier~1-only model achieves the highest per-design reward (3.77) but
covers only 7 topologies; 32 topologies with lower average validity is
the more useful configuration.

Signal circuits show the clearest GRPO advantage
(66$\pm$3\% sim-validity vs.\ 50$\pm$3\% REINFORCE). Power converters
remain hardest due to switching dynamics; GRPO still outperforms
alternatives. Full per-topology results are in
Appendix~Table~\ref{tab:per_topo_full}.

\subsection{Warm-Start Experiment}
\label{sec:warm-start}

A natural question is whether ARCS's fast inference can complement
search-based optimization. The \emph{warm-start} strategy works as follows:
ARCS generates 3 candidate designs, and the best seeds a GA population
of 20 individuals for 5 generations (113 SPICE simulations). The comparison
is cold-start GA (20 individuals, 10 generations, 220 simulations).

\begin{table}[t]
\centering
\caption{Warm-start comparison. GA-Warm uses ARCS-seeded initialization
with 49\% fewer SPICE simulations and 58\% less wall time.}
\label{tab:warmstart}
\begin{tabular}{@{}lccc@{}}
\toprule
Method & Reward & Sims & Time \\
\midrule
ARCS only     & 5.45 & 3   & 0.8\,s  \\
GA cold-start & 7.43 & 220 & 47.3\,s \\
GA warm-start & 7.18 & 113 & 19.7\,s \\
\bottomrule
\end{tabular}
\end{table}

Table~\ref{tab:warmstart} shows that on the 14 topologies where ARCS
produces valid initial designs (Wien bridge and Colpitts fall back to
cold-start), GA-Warm achieves \textbf{96.6\% of cold-start quality}
(7.18 vs.\ 7.43 reward) while using only 113 simulations, a 49\%
reduction. For signal-processing topologies, warm-start matches
cold-start exactly. On boost converter, warm-start
\emph{outperforms} cold-start (7.95 vs.\ 7.84), suggesting that
ARCS's learned initialization lands in a better basin of attraction.
The 58\% wall-clock speedup validates ARCS as a practical initializer
for search-based design workflows.

\subsection{Constrained Decoding}
\label{sec:constrained_results}

This section evaluates the grammar-constrained decoding scheme
(Section~\ref{sec:constrained}) across three constraint levels
(Grammar, Topology, Full) plus an unconstrained baseline (None).
To isolate constraints from model quality, the evaluation uses a
\emph{randomly initialized} model (no training),
demonstrating that the constraints alone suffice for structural
correctness.

\begin{table}[t]
\centering
\caption{Constrained decoding results (random-init model, 50 samples
across 32 topologies). Constrained decoding achieves 100\% structural
validity by construction and is faster than unconstrained sampling.}
\label{tab:constrained}
\begin{tabular}{@{}lcccc@{}}
\toprule
Level & Struct\,\% & Comp\,\% & Avg Comp & Time \\
\midrule
\textsc{None}     &  0.0\% &  0.0\% & ---  & 269\,ms \\
\textsc{Grammar}  & 100\%  &  0.0\% & 18.6 & 125\,ms \\
\textsc{Topology} & 100\%  & 100\%  & 4.3  & 27\,ms  \\
\textsc{Full}     & 100\%  & 100\%  & 4.3  & 25\,ms  \\
\bottomrule
\end{tabular}
\end{table}

All three constraint levels achieve 100\% structural validity,
compared to 0\% unconstrained, a \emph{formal guarantee} across all
32 topologies. \textsc{Topology} constraints additionally ensure 100\%
component type correctness. Surprisingly, constrained decoding
runs faster than unconstrained (25\,ms vs.\ 269\,ms): topology
constraints terminate sequences promptly instead of letting the model
generate meandering token streams. Since constraints are orthogonal to model quality (working
even with random initialization), they compose freely with RL fine-tuning.

\subsection{Inference-Time Compute Scaling}
\label{sec:bestofn}

ARCS generates a circuit in $\sim$30\,ms. Generating $N$ candidates
and selecting the best, without any SPICE simulation, trades minimal
additional inference cost for improved quality. The ranker uses mean
log-probability
(model confidence), which is computed for free during autoregressive
sampling. This is analogous to recent ``test-time compute
scaling''~\cite{snell2024scaling} results in language modeling,
where generating more candidates and selecting by a scoring function
yields substantial quality gains at low marginal cost.

\begin{table}[t]
\centering
\caption{Best-of-$N$ scaling with a trained model (80 specs, \textsc{Topology}
constraints). Model confidence (mean log-prob) improves monotonically;
SPICE reward peaks at $N\!=\!3$. Even $N\!=\!50$ costs only 1.6\,s, still
37$\times$ faster than random search.}
\label{tab:bestofn}
\begin{tabular}{@{}rccccc@{}}
\toprule
$N$ & Conf. & SimValid & Reward & Time & Speedup \\\midrule
1  & $-$1.16 & 72.5\% & 5.01 & 35\,ms  & 1,680$\times$ \\
3  & $-$0.82 & \textbf{85.0\%} & \textbf{5.48} & 97\,ms  & 606$\times$ \\
5  & $-$0.74 & 85.0\% & 5.40 & 164\,ms & 359$\times$ \\
10 & $-$0.67 & 77.5\% & 5.14 & 324\,ms & 181$\times$ \\
20 & $-$0.57 & 82.5\% & 5.18 & 641\,ms & 92$\times$ \\
50 & $-$0.52 & 82.5\% & 5.10 & 1.6\,s  & 37$\times$ \\
\midrule
RS & --- & 81.2\% & 7.28 & 58.8\,s & 1$\times$ \\
\bottomrule
\end{tabular}
\end{table}

SPICE reward peaks at $N\!=\!3$ (5.48, +9.3\% over single-shot)
then plateaus, even as model confidence continues improving, indicating
that log-probability is a useful but imperfect proxy for simulation quality.
Best-of-3 at 97\,ms remains 600$\times$ faster than random search (58.8\,s).

The non-monotonic reward curve at $N > 5$ motivates the learned
reward model (Section~\ref{sec:reward-model}) for more effective ranking.

\begin{figure}[t]
\centering
\begin{tikzpicture}
\begin{axis}[
  width=\columnwidth,
  height=4.8cm,
  xlabel={Number of candidates ($N$)},
  ylabel={SPICE Reward},
  xmode=log,
  log basis x=2,
  xtick={1,3,5,10,20,50},
  xticklabels={1,3,5,10,20,50},
  ymin=4.5, ymax=7.8,
  grid=major,
  grid style={dashed, gray!30},
  legend style={at={(0.02,0.98)}, anchor=north west, font=\scriptsize},
  mark options={solid},
]
\addplot[blue, thick, mark=*, mark size=2pt] coordinates
  {(1,5.01) (3,5.48) (5,5.40) (10,5.14) (20,5.18) (50,5.10)};
\addlegendentry{Best-of-$N$ (ARCS)}
\addplot[red, dashed, thick, domain=1:50] {7.28};
\addlegendentry{Random Search}
\addplot[orange, dashed, thick, domain=1:50] {7.48};
\addlegendentry{Genetic Algorithm}
\end{axis}
\end{tikzpicture}
\caption{Inference-time scaling curve. Best-of-$N$ with model confidence
ranking peaks at $N\!=\!3$ (reward 5.48, 97\,ms), narrowing the gap to
search baselines (RS 7.28 / 58.8\,s, GA 7.48 / 271\,s) at $>$600$\times$
less cost. The plateau beyond $N\!=\!5$ reflects the misalignment between
model confidence and SPICE reward.}
\label{fig:scaling}
\end{figure}
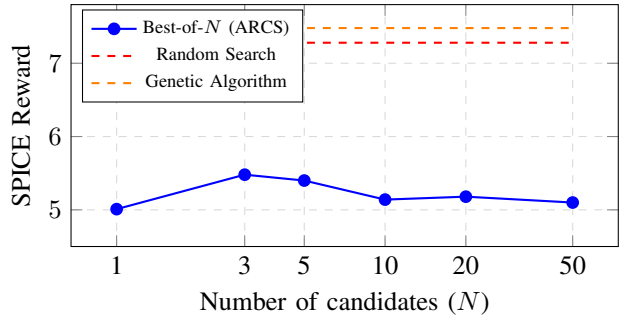

\subsection{Learned Reward Model}
\label{sec:reward-model}

To address the confidence--reward misalignment
(Section~\ref{sec:bestofn}), a 666K-parameter reward model (2-layer
bidirectional transformer encoder, 128-dim, 4 heads) is trained on
41K SPICE-simulated circuits using Huber loss. Token embeddings are
warm-started from the generator via SVD projection.

\begin{table}[t]
\centering
\caption{Learned reward model evaluation on 41K SPICE-simulated
circuits. The model achieves $r\!=\!0.988$ correlation with ground
truth and 93.4\% of predictions within $\pm$0.5 of the true reward.}
\label{tab:reward-model}
\begin{tabular}{@{}lc@{}}
\toprule
Metric & Value \\\midrule
Pearson correlation ($r$) & 0.988 \\
Spearman rank corr.\ ($\rho$) & 0.947 \\
MAE & 0.147 \\
RMSE & 0.390 \\
Within $\pm$0.5 & 93.4\% \\
Within $\pm$1.0 & 97.8\% \\
\bottomrule
\end{tabular}
\end{table}

The reward model achieves $r\!=\!0.988$ correlation with ground-truth
reward (Table~\ref{tab:reward-model}).

\textbf{Ranking comparison.} A comparison of confidence-based and
reward-model-based ranking for Best-of-$N$ candidate selection,
evaluating the selected circuit via SPICE simulation (16 test
specs, all topologies).

\begin{table}[t]
\centering
\caption{Best-of-$N$ ranking comparison: confidence vs.\ learned reward
model (80 trials per cell, 5 seeds $\times$ 16 test specs).
The reward model consistently improves ranking at $N\!=\!3$
($+2.0$--$2.1\%$) for both SL and RL generators.}
\label{tab:ranking-compare}
\begin{tabular}{@{}r ccc ccc@{}}
\toprule
     & \multicolumn{3}{c}{\textbf{SL Generator}}
     & \multicolumn{3}{c}{\textbf{RL Generator}} \\
\cmidrule(lr){2-4}\cmidrule(lr){5-7}
$N$  & Conf & RM & $\Delta$\%
     & Conf & RM & $\Delta$\% \\\midrule
1    & 5.16 & 5.16 & ---
     & 5.06 & 5.06 & --- \\
3    & 5.05 & 5.15 & \textbf{+2.0}
     & 5.08 & 5.18 & \textbf{+2.1} \\
5    & 5.03 & 4.71 & $-$6.5
     & 5.12 & 5.16 & +0.7 \\
10   & 5.08 & 5.22 & +2.7
     & 5.30 & 5.16 & $-$2.5 \\
20   & 5.18 & 4.81 & $-$7.2
     & 5.04 & 5.17 & +2.5 \\
\bottomrule
\end{tabular}
\end{table}

At $N\!=\!3$, the reward model consistently outperforms confidence
ranking for both SL and RL generators ($+2.0\%$ and $+2.1\%$),
confirming that the learned reward signal captures quality aspects
that log-probability misses. The reward model is most reliably
valuable at moderate $N$ ($\approx$3), providing a consistent,
training-free quality boost at negligible cost.

\section{Discussion}
\label{sec:discussion}

\textbf{Architecture vs.\ RL vs.\ GRPO.} Architectural changes
(Baseline$\to$GT) yield $+0.49$ reward from inductive
bias alone. REINFORCE degrades validity ($-1.9$\,pp) due to
cross-topology reward mismatch, while GRPO improves it ($+7.7$\,pp
over SL) with 10$\times$ fewer steps. Per-topology advantage
normalization is essential when training spans circuit families with
heterogeneous reward scales.

\textbf{Amortized vs.\ iterative design.} ARCS and search baselines
are complementary: ARCS-seeded GA recovers 96.6\% of cold-start
quality with 49\% fewer simulations (Section~\ref{sec:warm-start}).

\textbf{Structure by grammar, semantics by learning.}
Constrained decoding guarantees 100\% structural validity by
construction; GRPO optimizes \emph{value quality} within the valid
subspace. This decomposition mirrors the separation of syntax and
semantics in programming language design~\cite{hokamp2017lexically}.

\textbf{Scalability.} ARCS scales to new topologies by writing a
single SPICE template, with no manual curation needed. Extending to
transistor-level IC circuits requires more templates and likely
larger models.

\textbf{Limitations.}
\begin{itemize}
    \item Per-design quality remains below search baselines (5.48
    Best-of-3 vs.\ 7.48 GA). Scaling to 50--100M parameters with more
    training data is the primary direction for improvement.
    \item While constrained decoding guarantees \emph{structural}
    validity, \emph{simulation} validity (i.e., whether SPICE produces
    correct output) still depends on learned component values.
    \item The value tokenizer's 500-bin resolution ($\sim$28 bins/decade)
    limits precision to $\sim$3.5\% relative error, which may be
    insufficient for sensitive RF or precision analog circuits.
\end{itemize}

\textbf{Future work.}
Scaling to 50--100M parameters, curriculum learning over topology
difficulty, end-to-end CCFM fine-tuning with SPICE-in-the-loop, and
extending constraints to enforce Kirchhoff's laws.

\textbf{Code and data availability.} Code, data, and trained checkpoints
are available at \url{https://github.com/tusharpathaknyu/ARCS}.

\section{Conclusion}
\label{sec:conclusion}

ARCS generates complete, SPICE-simulatable analog circuits conditioned
on target specifications across 32 topologies, in milliseconds rather
than minutes. Three findings stand out. First, per-topology advantage
normalization (GRPO) resolves the cross-topology reward mismatch that
causes REINFORCE to regress, improving simulation validity by +9.6\,pp
with 10$\times$ fewer RL steps. Second, grammar-constrained decoding
guarantees 100\% structural validity by construction, reaching 85\%
simulation validity in 97\,ms with Best-of-3 selection. Third, a hybrid
pipeline combining complementary generators with SPICE-based ranking
achieves 99.9\% simulation validity using only 8 SPICE
evaluations, 40$\times$ fewer than genetic algorithms.

Per-design quality remains below search baselines (5.48 vs.\ 7.48
reward). But ARCS is $>$1000$\times$ faster, and ARCS-seeded GA
recovers 96.6\% of cold-start quality with 49\% fewer simulations.
The two paradigms are complementary. Amortized circuit generation is
not a replacement for search. It is a practical starting point.

\appendix
\section{Per-Topology Results}
\label{app:per_topology}

Table~\ref{tab:per_topo_full} presents hybrid-ranked results
for all 32 evaluated topologies (50 samples each, bootstrap 95\% CI),
sorted by mean reward.

\begin{table}[h]
\centering
\caption{Per-topology hybrid-ranked results (50 samples, spec-aware reward).}
\label{tab:per_topo_full}
\small
\begin{tabular}{@{}llcc@{}}
\toprule
Topology & Cat & Reward & 95\% CI \\
\midrule
current\_mirror      & B & 8.00 & [8.00, 8.00] \\
voltage\_doubler     & P & 7.96 & [7.96, 7.97] \\
shunt\_regulator     & R & 7.77 & [7.64, 7.85] \\
series\_regulator    & R & 7.59 & [7.41, 7.73] \\
sallen\_key\_lowpass & F & 7.49 & [7.36, 7.60] \\
common\_emitter     & B & 7.05 & [6.90, 7.20] \\
wien\_bridge         & O & 7.00 & [7.00, 7.00] \\
colpitts            & O & 7.00 & [7.00, 7.00] \\
common\_collector   & B & 7.00 & [7.00, 7.00] \\
hartley             & O & 7.00 & [7.00, 7.00] \\
\midrule
inverting\_amp       & A & 6.81 & [6.76, 6.86] \\
phase\_shift         & O & 6.80 & [6.50, 7.00] \\
common\_base        & B & 6.67 & [6.56, 6.80] \\
differential\_amp   & A & 6.63 & [6.54, 6.73] \\
inv\_summing\_amp   & A & 6.60 & [6.50, 6.68] \\
half\_bridge        & P & 6.52 & [6.31, 6.72] \\
cuk                  & P & 6.49 & [6.26, 6.72] \\
cascode              & B & 6.42 & [6.31, 6.53] \\
charge\_pump         & P & 6.41 & [6.40, 6.42] \\
buck                 & P & 6.37 & [6.25, 6.49] \\
\midrule
noninverting\_amp    & A & 6.30 & [6.14, 6.44] \\
push\_pull           & P & 6.04 & [5.77, 6.32] \\
boost                & P & 6.01 & [5.85, 6.17] \\
transimpedance\_amp & A & 6.00 & [6.00, 6.00] \\
twin\_t\_notch       & F & 5.76 & [5.54, 5.98] \\
zeta\_converter     & P & 5.71 & [5.51, 5.93] \\
instrum\_amp        & A & 5.40 & [5.24, 5.58] \\
buck\_boost          & P & 5.34 & [5.22, 5.48] \\
sepic                & P & 5.24 & [4.99, 5.50] \\
sk\_highpass         & F & 5.00 & [5.00, 5.00] \\
sk\_bandpass         & F & 4.99 & [4.91, 5.06] \\
state\_var\_filter   & F & 4.45 & [4.30, 4.63] \\
\bottomrule
\end{tabular}
\end{table}

Current mirrors and voltage doublers achieve near-perfect scores (8.0)
due to simple operating conditions. The weakest topologies, state variable
filter (4.45) and Sallen-Key bandpass/highpass (5.0), reflect the
difficulty of precise frequency control in multi-pole filters.

\bibliographystyle{IEEEtran}

\end{document}